\documentclass{article}
\usepackage[preprint]{neurips_2026}

\usepackage{amsmath,amssymb,amsfonts,amsthm,mathtools}
\usepackage[utf8]{inputenc}
\usepackage[T1]{fontenc}
\usepackage{hyperref}
\usepackage{url}
\usepackage{booktabs}
\usepackage{nicefrac}
\usepackage{microtype}
\usepackage{xcolor}
\usepackage{graphicx}
\usepackage{subcaption}
\usepackage{multirow}
\usepackage{algorithm}
\usepackage{algorithmic}
\usepackage[capitalize,noabbrev]{cleveref}

\theoremstyle{plain}

\theoremstyle{definition}

\theoremstyle{remark}

\newcommand{\pairmem}{e^{\mathrm{pair}}}
\newcommand{\tripmem}{e^{\mathrm{tri}}}
\newcommand{\edgefeat}{x^{\mathrm{edge}}}

\newcommand{\bucket}{\mathrm{bucket}}

\microtypesetup{expansion=false}

\title{PaMM: Periodic Motif Memory for Atomistic Models with an Explicit Local-Structure Interface}

\author{
Ryan Dong \\
Independent Research \\
\texttt{zwdong618@gmail.com}
}
\date{}

\begin{document}

\maketitle

\begin{abstract}
Periodic crystals repeatedly instantiate similar local coordination motifs across translated cells and chemically related structures, but current equivariant atomistic models usually encode these patterns only implicitly in dense edge features. We introduce PaMM, a periodic motif memory that augments the UMA eSCN-MD edge encoder with explicit pair and triplet lookup features. Pair motifs are keyed by $(Z_j, Z_i, b_r)$ and triplet motifs by $(Z_j, Z_i, Z_k, b_\theta)$, hashed into fixed-size tables and fused with the baseline edge representation through lightweight gate-only and affine-equipped variants.

We evaluate PaMM in a matched UMA-S OMAT setting and focus on a narrow question: whether explicit motif memory helps at a fixed intermediate training budget. At the 10k-step checkpoint, both PaMM variants improve over the plain baseline; gate-only gives the best energy MAE, while the affine-equipped variant gives the best force MAE. A matched 20k follow-up keeps the same operating-point picture. Aligned controls show that the gain weakens for pair-only, triplet-only, random-bucket, and parameter-matched MLP alternatives, suggesting that the benefit is tied to structured pair/triplet organization rather than generic added capacity. A within-OMAT24 source-family evaluation also shows small but consistent gains across held-out generation families.

We therefore make a focused claim: in the studied UMA-S + OMAT regime, explicit pair/triplet motif memory is a useful inductive bias for periodic atomistic modeling. We do not claim broad cross-dataset transfer, a uniquely preferred fusion variant, or strong scientific interpretability beyond a more inspectable local-structure interface.

\end{abstract}

\section{Introduction}

Large pretrained atomistic models are starting to cover molecules, materials, catalysts, and related chemical domains within a shared backbone, with UMA as a recent example \citep{uma2025}. In periodic materials, however, much of the predictive signal is strongly repetitive: similar pair distances, coordination angles, and local environments recur across translated unit cells, neighboring coordination shells, and chemically related crystals. Modern equivariant backbones such as NequIP, MACE, and EquiformerV2 can in principle represent these patterns well, but they usually do so implicitly through continuous edge features and message passing \citep{nequip2022,mace2022,equiformerv2}. The same local geometry must therefore be reconstructed repeatedly inside dense hidden states rather than exposed as reusable model state.

This distinction matters most when one wants controlled additional capacity rather than a different backbone. Scaling width, depth, or dataset mixture changes many ingredients at once, whereas an explicit memory path can target a more specific hypothesis: recurring local chemistry may be worth storing in a bounded representation that the host network can read out when needed. In periodic materials, a natural choice is to key the memory by low-order geometric objects that recur frequently and are already central to atomistic reasoning, namely pairwise distances and local triplet angles.

We therefore introduce \emph{PaMM}, a periodic motif memory attached to the UMA eSCN-MD edge encoder. PaMM retrieves a pair memory from element-pair and distance keys and a triplet memory from local angular configurations. The retrieved features are fused back into the baseline edge representation through lightweight gating, with an optional additional affine modulation path inside the message-passing stack. The downstream equivariant operator is left unchanged. In this sense, PaMM is an additive mechanism rather than an alternative backbone: it contributes explicit local-structure state without replacing the host's continuous geometric pipeline.

The empirical scope of the paper is intentionally focused. We study PaMM inside the UMA-S eSCN-MD host on OMAT and ask whether explicit motif memory helps under a matched intermediate training budget. The main text is organized around four concrete questions. First, does the memory path improve the plain host at locked 10k and 20k checkpoints? Second, are the observed gains stable enough under reseeding to support the core claim? Third, do aligned controls indicate that the effect comes from structured pair/triplet organization rather than from generic extra parameters? Fourth, how does the method behave when we vary the memory size or regroup evaluation by held-out OMAT source family?

Our conclusions are narrow by design. At the locked 10k checkpoint, both PaMM variants improve over the plain baseline; the gate-only variant is strongest on energy, while the affine-equipped variant is strongest on force. A matched 20k follow-up preserves this picture. Additional controls show that the gain weakens when the memory is reduced to pair-only or triplet-only lookup, when structured keys are replaced by random buckets, or when the structured branch is replaced by a parameter-matched MLP. A bucket sweep indicates diminishing returns beyond the mid-sized regime, and a source-family evaluation within OMAT24 shows small but consistent gains across held-out generation families. We therefore position PaMM as a useful inductive bias for the studied host/domain/budget regime, not as a claim of universal backbone transfer, clean optimizer-independence, or strong chemistry-level interpretability.

The contributions of the paper are:
\begin{enumerate}
\item We introduce an explicit pair/triplet motif memory for periodic atomistic modeling, implemented as fixed-size hash lookup over local geometric keys rather than as another dense message-passing stage.
\item We integrate this memory into UMA eSCN-MD through two lightweight readout mechanisms, a gate-only variant and an affine-equipped variant, while leaving the downstream equivariant operator unchanged.
\item We provide a focused evidence package in the UMA-S + OMAT setting that covers matched 10k and 20k checkpoints, three-seed robustness at the main 10k budget, aligned controls, and held-out source-family evaluation.
\item We analyze how the method uses its added capacity through training-dynamics and bucket-count studies, which helps distinguish useful explicit memory from generic parameter growth.
\end{enumerate}

\section{Related Work}
\label{sec:related}

Recent atomistic ML has moved from narrow task-specific models toward broader pretrained systems such as UMA, while parallel work in the Open Catalyst ecosystem has established the value of multi-domain pretraining and transfer for atomistic prediction \citep{uma2025,kolluru2022transfer,shoghi2023molecules,lan2023adsorbml,chanussot-2021-open-catal,tran-2023-open-catal}. PaMM operates in this regime but changes a different axis. Instead of enlarging the backbone or changing the pretraining mixture, it adds a bounded explicit memory path for recurring local geometry.

Equivariant interatomic networks such as NequIP, MACE, and EquiformerV2 improve accuracy mainly through stronger continuous geometric operators \citep{nequip2022,mace2022,equiformerv2}. Higher-order structure is usually represented through angular bases, tensor products, or richer higher-body interactions inside the message-passing stack. PaMM is complementary rather than competitive with this line of work. It leaves the downstream equivariant operator intact and instead augments the scalar edge representation seen by that operator.

This difference in placement is central. In existing atomistic backbones, pairwise and angular regularities influence prediction indirectly through learned continuous features. PaMM instead treats a subset of these regularities as explicit reusable state, accessed through fixed-capacity lookup tables. The goal is not to discretize the whole geometric computation, but to expose recurring low-order motifs in a form that can be reused across many edges while the host backbone continues to handle directional and equivariant processing.

The closest intuition comes from recent work on scalable conditional memory in language modeling, where repeated patterns are handled by targeted lookup rather than by uniform dense computation \citep{engramllm2026}. Our setting is substantially different: periodic geometry is local, structured, and constrained by atomic neighborhoods rather than by token sequences. The novelty here is therefore not generic hashing inside an atomistic model. It is the decision to represent recurring pair and triplet coordination motifs as explicit model state attached to the edge encoder of a modern pretrained atomistic backbone.

\section{Method}
\label{sec:method}

We modify the UMA eSCN-MD edge encoder and keep the downstream equivariant message-passing stack unchanged. Let a periodic structure induce a directed neighborhood graph whose edges $(j \rightarrow i)$ connect neighbor atoms within the host cutoff. In the plain UMA encoder, each directed edge is represented by scalar features derived from the pair distance and the source/target atom types. PaMM augments this scalar edge description with two additional retrieved features: a pair memory and a triplet memory. The central design choice is to expose recurring local motifs through explicit lookup while leaving the host's geometric operator intact.

\begin{figure*}[t]
\centering
\includegraphics[width=0.96\linewidth]{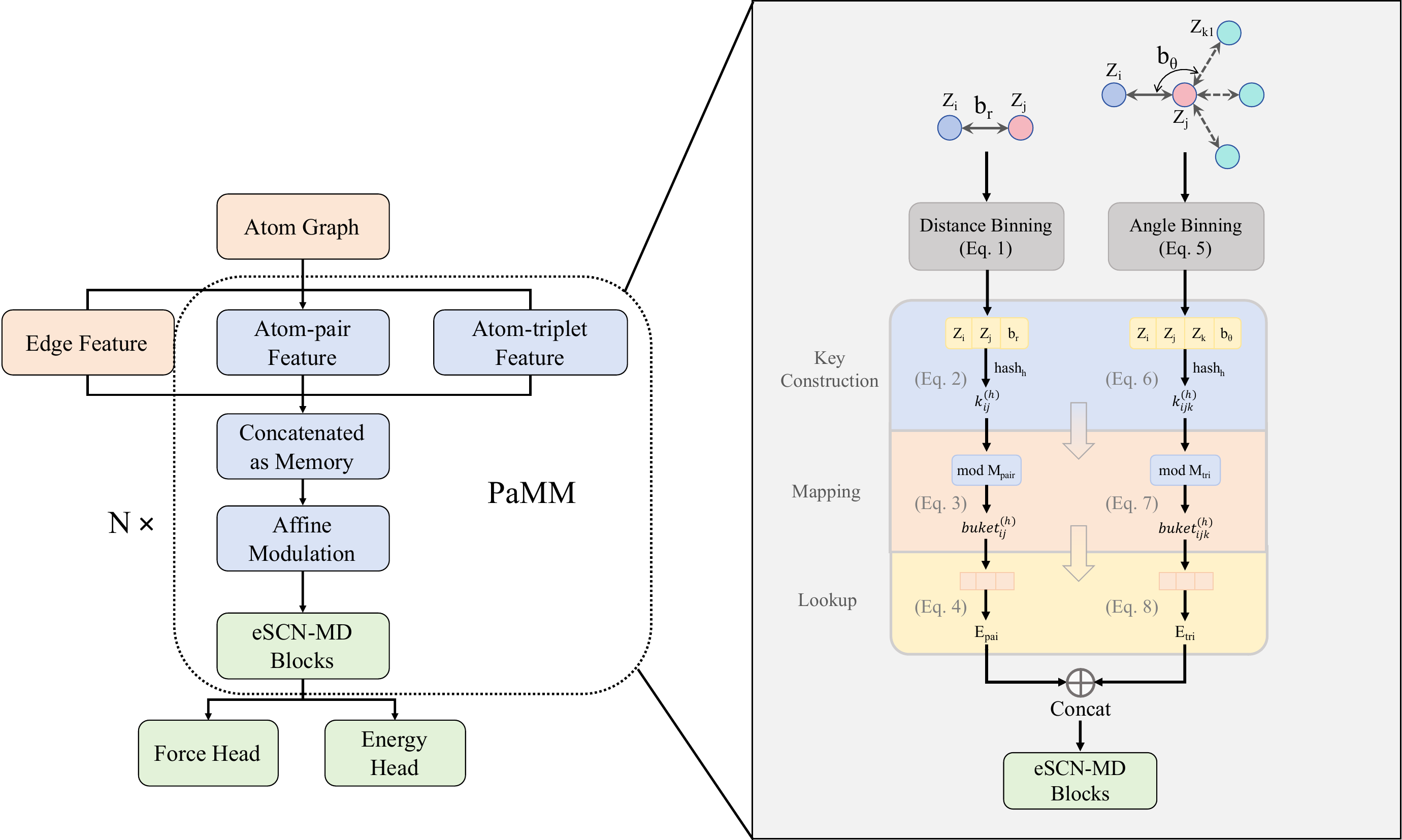}
\caption{\textbf{PaMM augments UMA eSCN-MD with an explicit motif interface.} Pair motifs are keyed by atom types and discretized distances, while triplet motifs are keyed by local angular configurations. The retrieved pair/triplet memories are concatenated with the baseline edge representation and then consumed by lightweight fusion modules. In the affine-equipped variant, the same memory also drives a per-layer modulation path inside the eSCN-MD stack.}
\label{fig:method_overview}
\end{figure*}

\subsection{Motif Lookup Construction}

The two lookup paths are designed to capture complementary local information. The pair memory tracks recurring element-pair and distance patterns at the level of a single edge. The triplet memory adds coarse angular context around the center atom, allowing the encoder to distinguish edges that share a similar pairwise geometry but belong to different local coordination environments.

For edge $(j \rightarrow i)$ with atomic numbers $(Z_j, Z_i)$ and distance $r_{ij}$, we quantize the distance into a bin
\begin{equation}
b_r(i,j) = \left\lfloor \frac{r_{ij}}{r_{\max}} B_r \right\rfloor.
\end{equation}
For hash function index $h \in \{1,\dots,H\}$, we form a pair key
\begin{equation}
k^{(h)}_{ij} = Z_j (p+h)^2 + Z_i (p+h) + b_r(i,j),
\end{equation}
map it into a bucket $\bucket^{(h)}_{ij} = k^{(h)}_{ij} \bmod M_{\mathrm{pair}}$, and retrieve
\begin{equation}
\pairmem_{ij} = \frac{1}{H}\sum_{h=1}^{H} E^{(h)}_{\mathrm{pair}}\!\left(\bucket^{(h)}_{ij}\right).
\end{equation}
This lookup deliberately compresses geometry into a reusable low-order signature. Exact directional information is not discarded from the full model, because the downstream eSCN-MD stack still receives the host's continuous geometric inputs; PaMM only adds a bounded summary of recurring pair structure.

To encode angular context, we enumerate triplets $(j,i,k)$ around the center atom $i$, quantize the angle $\theta_{jik}$ into $b_\theta(j,i,k)$, and hash
\begin{equation}
k^{(h)}_{jik} = Z_j (p+h)^3 + Z_i (p+h)^2 + Z_k (p+h) + b_\theta(j,i,k).
\end{equation}
The triplet lookup is
\begin{equation}
\tripmem_{ij} =
\frac{1}{|\mathcal{T}_{ij}|}
\sum_{(j,i,k)\in\mathcal{T}_{ij}}
\frac{1}{H}\sum_{h=1}^{H}
E^{(h)}_{\mathrm{tri}}\!\left(k^{(h)}_{jik} \bmod M_{\mathrm{tri}}\right),
\end{equation}
where $\mathcal{T}_{ij}$ is the set of triplets that contribute to edge $(j \rightarrow i)$.
The triplet aggregation is edge-conditioned but center-local: every contributing triplet shares the same central atom $i$ and edge endpoint $j$, so the resulting memory summarizes the angular environment relevant to that directed interaction.

The two memories are concatenated with the baseline edge feature:
\begin{equation}
\edgefeat_{ij} =
[\phi_r(r_{ij}); \phi_s(Z_j); \phi_t(Z_i); \pairmem_{ij}; \tripmem_{ij}].
\end{equation}
This construction exposes recurring pair and local-angle motifs as explicit, bounded-capacity model state. The bucket counts $M_{\mathrm{pair}}$ and $M_{\mathrm{tri}}$ are therefore direct memory-capacity knobs: increasing them reduces collisions and increases representational resolution, while keeping the lookup path compact and localized.

\subsection{Memory Readout}

We study two ways of consuming the same memory. The \emph{gate-only} variant applies a bounded scalar gate to the augmented edge feature. The \emph{affine-equipped} variant keeps the same edge-level path and additionally injects memory-conditioned channel-wise affine modulation inside each eSCN-MD layer. In both cases, the downstream equivariant operator is unchanged; PaMM only changes how local structure is represented before or alongside that operator.

\paragraph{Edge-gated readout.}
Let $\edgefeat_{ij}$ denote the augmented edge feature from the pair/triplet lookup. The gate-only variant computes
\begin{equation}
g_{ij}=\mathrm{clip}\!\Bigl(1+\lambda\,\mathrm{SiLU}\bigl(\alpha \langle W_q\edgefeat_{ij}, W_k\edgefeat_{ij}\rangle\bigr),\, g_{\min}, g_{\max}\Bigr),
\end{equation}
and rescales the edge feature as
\begin{equation}
\tilde{\edgefeat}_{ij}=\edgefeat_{ij}\odot g_{ij}.
\end{equation}
This is an edge-local scalar modulation rather than a multi-edge attention mechanism. The readout remains strictly local: each edge is reweighted only by its own retrieved motif state.

\paragraph{Layer-wise affine modulation.}
Let $m_{ij}=[\pairmem_{ij};\tripmem_{ij}]$. For each message-passing layer $\ell$, the affine-equipped variant predicts a memory-conditioned scale and bias
\begin{equation}
(\Delta g_{ij}^{(\ell)},\Delta b_{ij}^{(\ell)})=\mathrm{MLP}_{\ell}(m_{ij}),
\end{equation}
and updates the layer-specific edge state through
\begin{equation}
e_{ij}^{(\ell)} \leftarrow e_{ij}^{(\ell)}\odot\bigl(1+\alpha_{\mathrm{aff}}\tanh(\Delta g_{ij}^{(\ell)})\bigr)+\beta_{\mathrm{aff}}\Delta b_{ij}^{(\ell)}.
\end{equation}
The gate-only and affine-equipped variants therefore share the same explicit motif memory; they differ only in whether that memory is used as an edge-level gate alone or also as an internal per-layer modulation signal.

\subsection{Design Choices and Cost}

PaMM is intended to be an additive mechanism rather than a second interaction stack. The pair lookup targets the most repeated local regularity, namely element-pair and distance patterns. The triplet lookup adds a coarse notion of local coordination without introducing higher-order tensor operations into the host backbone. The two readout variants then test whether this explicit state is most useful as a single pre-message-passing gate or as a signal that should also modulate deeper edge channels.

\begin{table}[t]
\centering
\caption{\textbf{Added cost of PaMM relative to the host edge encoder.} Here $E$ is the number of directed edges, $T$ is the number of enumerated triplets, $d$ is the memory width, $H$ is the number of hash functions, and $L$ is the number of message-passing layers.}
\label{tab:method_cost}
\small
\begin{tabular}{p{0.30\linewidth}p{0.60\linewidth}}
\toprule
Component & Added cost \\
\midrule
Pair memory & parameters $\mathcal{O}(H M_{\mathrm{pair}} d)$; lookup work linear in $E$ \\
Triplet memory & parameters $\mathcal{O}(H M_{\mathrm{tri}} d)$; enumeration and aggregation linear in $T$ \\
Gated readout & edge-local projections and scalar modulation, linear in active edges \\
Affine modulation & per-layer memory MLP plus channel-wise affine update, linear in $L$, $E$, and edge width \\
\bottomrule
\end{tabular}
\end{table}

The resulting overhead remains local. PaMM does not add another message-passing stack and does not introduce all-to-all interactions across atoms. Instead, it adds lookup and modulation costs that scale with the same edge and triplet objects already present in the host neighborhood construction. Because the added state is an index-addressed lookup table rather than a dense equivariant activation block, it is also deployment-flexible: at the paper default it occupies only tens of MB and can in principle be kept off the critical GPU-resident path if VRAM is the bottleneck. In the implemented recipe, the lookup and gate submodules are also placed in a reduced learning-rate optimizer group, because this was the stable setting available in code. Accordingly, the paper evaluates the complete PaMM recipe rather than claiming a perfectly isolated architecture-only intervention.

\section{Experiments}
\label{sec:experiments}

\subsection{Setup}

We evaluate PaMM as a modification to the UMA-S-1p1 eSCN-MD host. All main-text experiments use the same OMAT local split of \texttt{dataset=uma}, the same training entrypoint, and matched optimization settings. Unless otherwise noted, PaMM uses separate pair and triplet tables, two hash functions, discretized distance and angle bins, and the reduced lookup/gate learning-rate group described in \Cref{sec:method}. We report energy and force mean absolute error (MAE), with lower values indicating better performance.

The main paper is built around matched-budget comparisons rather than around a best-of-many-run search. We focus first on a locked 10k-step checkpoint, where the central question is whether explicit motif memory helps before full convergence. We then add a matched 20k follow-up to test whether the same mechanism remains competitive at a slightly longer horizon. The remainder of the main section asks whether the effect survives reseeding, whether it depends on structured motif organization, whether it is visible across held-out OMAT source families, and how it changes as the memory capacity is varied.

\paragraph{Variant semantics.}
The \emph{plain baseline} disables PaMM entirely. The \emph{gate-only PaMM} variant keeps structured pair/triplet lookup and applies only the scalar edge gate from \Cref{sec:method}. The \emph{affine-equipped PaMM} variant keeps the same structured lookup and edge-level path but additionally injects the memory-conditioned affine modulation inside each eSCN-MD layer. For the aligned controls, \emph{pair-only} and \emph{triplet-only} remove one motif source, \emph{no-gate} keeps structured lookup while bypassing the scalar gate, \emph{random-bucket} preserves lookup-style capacity but destroys meaningful geometric keys, and the \emph{MLP control} replaces structured lookup by a parameter-matched generic learned branch.

\subsection{Matched-Budget Results}

\begin{table*}[t]
\centering
\caption{\textbf{OMAT results at matched 10k and 20k budgets.} Lower is better for all MAEs.}
\label{tab:main_budget}
\small
\setlength{\tabcolsep}{4pt}
\begin{tabular}{lcccccccc}
\toprule
& \multicolumn{4}{c}{10k checkpoint} & \multicolumn{4}{c}{20k checkpoint} \\
\cmidrule(lr){2-5} \cmidrule(lr){6-9}
Variant & Val E & Val F & Test E & Test F & Val E & Val F & Test E & Test F \\
\midrule
Plain baseline & 0.3516 & 0.0939 & 0.4772 & 0.0980 & 0.2667 & 0.0830 & 0.3866 & 0.0869 \\
Gate-only PaMM & \textbf{0.3269} & 0.0926 & \textbf{0.4470} & 0.0964 & \textbf{0.2638} & 0.0829 & 0.3808 & 0.0867 \\
Affine-equipped PaMM & 0.3312 & \textbf{0.0924} & 0.4612 & \textbf{0.0961} & 0.2696 & \textbf{0.0828} & \textbf{0.3757} & \textbf{0.0865} \\
\bottomrule
\end{tabular}
\end{table*}

At the 10k checkpoint, both PaMM variants improve over the plain baseline. Gate-only gives the best validation and test energy MAE, while the affine-equipped variant gives the best validation and test force MAE. This is the main result of the paper: explicit pair/triplet memory helps in the studied matched-budget regime.

The 20k follow-up keeps the same broader picture but sharpens the interpretation. Structured memory remains competitive, yet the preferred operating point depends on how the memory is used. Gate-only remains best on validation energy, while the affine-equipped variant is strongest on force and test energy. We therefore do not frame the paper as selecting a single universally best fusion rule; the central claim is about the explicit memory path itself.

\begin{figure*}[t]
\centering
\includegraphics[width=0.88\linewidth]{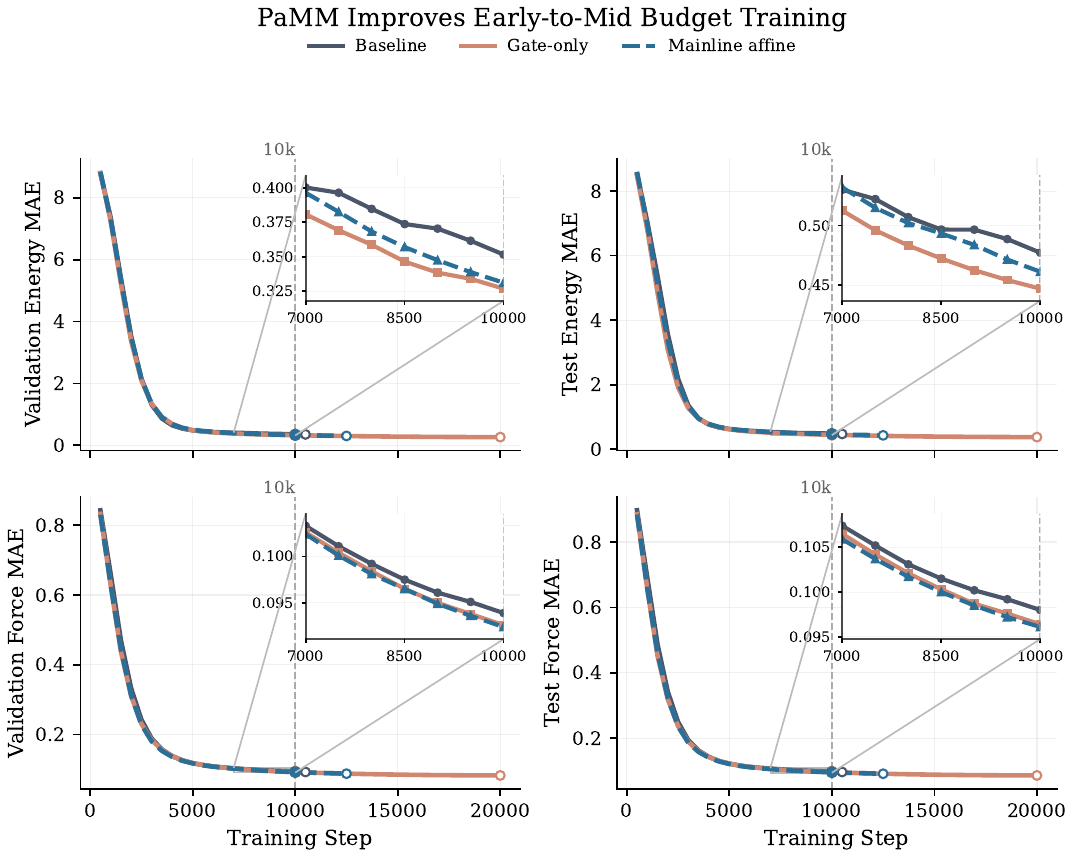}
\caption{\textbf{Anytime validation and test curves under the matched protocol.} Both PaMM variants improve on the plain UMA-S baseline before full convergence. Gate-only is strongest on energy near the locked budget, while the affine-equipped variant is strongest on force.}
\label{fig:main_anytime_main}
\end{figure*}

\paragraph{Training dynamics.}
\Cref{fig:main_anytime_main} shows that the improvement is not confined to a single reporting checkpoint. The gate-only variant pulls ahead on energy early and remains favorable around the locked 10k budget, while the affine-equipped variant more consistently favors force. This matters because PaMM is intended as a controlled additional mechanism rather than as a late-training rescue effect. The figure therefore supports the interpretation that explicit motif memory changes the optimization trajectory in a useful way during the main training horizon studied in this paper.

\subsection{Seed Stability at 10k}

\begin{table*}[t]
\centering
\caption{\textbf{Three-seed robustness summary at the 10k checkpoint.} We report mean $\pm$ std over three seeds for the same host and protocol as the main comparison. Lower is better for all MAEs.}
\label{tab:seed10k}
\scriptsize
\setlength{\tabcolsep}{4pt}
\begin{tabular}{lcccc}
\toprule
Variant & Val E & Val F & Test E & Test F \\
\midrule
Plain baseline & 0.3229$\pm$0.0012 & 0.0951$\pm$0.0004 & 0.4467$\pm$0.0094 & 0.0984$\pm$0.0005 \\
Gate-only PaMM & \textbf{0.3193}$\pm$0.0026 & \textbf{0.0951}$\pm$0.0001 & \textbf{0.4440}$\pm$0.0037 & \textbf{0.0984}$\pm$0.0001 \\
Affine-equipped PaMM & 0.3243$\pm$0.0018 & 0.0953$\pm$0.0004 & 0.4474$\pm$0.0069 & 0.0984$\pm$0.0004 \\
\bottomrule
\end{tabular}
\end{table*}

\Cref{tab:seed10k} shows that the energy-oriented gate-only advantage is preserved on average across three runs, while the force differences remain extremely small. We therefore use the single exported 10k checkpoints in \Cref{tab:main_budget} as the clean operating-point comparison, but read them together with this robustness summary: the gains are real enough to survive reseeding, yet not large enough to justify an overly broad claim about uniformly dominant performance. In other words, the method appears directionally stable for the paper's main claim, but the absolute margin is still modest and metric-dependent.

\subsection{Aligned Controls}

\begin{table}[t]
\centering
\caption{\textbf{Aligned control experiments at the 10k checkpoint.} Lower is better for all MAEs.}
\label{tab:controls10k}
\small
\setlength{\tabcolsep}{4pt}
\begin{tabular}{lcccc}
\toprule
Variant & Val E & Val F & Test E & Test F \\
\midrule
No-gate & \textbf{0.3317} & \textbf{0.0939} & \textbf{0.4601} & \textbf{0.0978} \\
Pair-only & 0.3407 & 0.0952 & 0.4673 & 0.0990 \\
Triplet-only & 0.3400 & 0.0952 & 0.4627 & 0.0988 \\
Random-bucket & 0.3402 & 0.0953 & 0.4623 & 0.0990 \\
MLP control & 0.3595 & 0.0979 & 0.5089 & 0.1020 \\
\bottomrule
\end{tabular}
\end{table}

\Cref{tab:controls10k} shows that the gain is structure-sensitive. Removing the gate does not erase the benefit of the full pair+triplet memory, so lookup itself matters. Pair-only and triplet-only variants are both weaker than the full memory path, suggesting that the two motif types are complementary. Random buckets weaken performance relative to structured keys, and a parameter-matched MLP control is worse still. Together these controls argue against the explanation that PaMM wins merely by adding generic capacity. They also clarify the role of the two readout choices in \Cref{sec:method}: the useful signal is already present in the structured lookup itself, while the gate and affine branches mainly determine how strongly and where that signal is injected into the host.

\subsection{Source-Family Evaluation}

\begin{table}[t]
\centering
\caption{\textbf{Held-out OMAT24 source-family evaluation at the 10k checkpoint.} Lower is better.}
\label{tab:omat-family}
\small
\setlength{\tabcolsep}{3.5pt}
\begin{tabular}{lcccc}
\toprule
Family & Base E & Base F & PaMM E & PaMM F \\
\midrule
AIMD-1000 & 0.0211 & 0.0596 & \textbf{0.0198} & \textbf{0.0588} \\
AIMD-3000 & 0.0168 & 0.1042 & \textbf{0.0163} & \textbf{0.1033} \\
Rattled & 0.0242 & 0.1138 & \textbf{0.0226} & \textbf{0.1109} \\
\shortstack[l]{Rattled-\\subsampled} & 0.0224 & 0.1249 & \textbf{0.0220} & \textbf{0.1224} \\
Rattled-relax & 0.0422 & 0.1614 & \textbf{0.0420} & \textbf{0.1612} \\
\midrule
Overall & 0.0235 & 0.1096 & \textbf{0.0226} & \textbf{0.1078} \\
\bottomrule
\end{tabular}
\end{table}

To test whether the gain is limited to a single pooled split, we also regroup OMAT24 evaluation by held-out source family. The evaluated PaMM checkpoint is slightly better than the plain baseline on both energy and force in all five families and on the overall aggregate. This is still a within-OMAT24 check rather than cross-dataset transfer, but it shows that the improvement is not confined to one pooled validation slice. The result is consistent with the motivating intuition of PaMM: recurring local motifs appear across multiple data-generation families, so an explicit motif interface should help beyond one particular validation shard.

\subsection{Memory Capacity Sweep}

\begin{table}[t]
\centering
\caption{\textbf{Bucket-count sweep for PaMM at the matched 10k-step budget.} Pair and triplet tables use the same bucket count. Lower is better for all MAEs.}
\label{tab:bucket_sweep_main}
\small
\begin{tabular}{lcccc}
\toprule
Buckets & Val E & Val F & Test E & Test F \\
\midrule
2048 & 0.04435 & 0.10507 & 0.06121 & 0.10851 \\
4096 & 0.04373 & 0.10515 & 0.06009 & 0.10844 \\
8192 & \textbf{0.04301} & \textbf{0.10464} & 0.05905 & \textbf{0.10833} \\
16384 & 0.04315 & 0.10487 & \textbf{0.05818} & 0.10849 \\
\bottomrule
\end{tabular}
\end{table}

The bucket sweep in \Cref{tab:bucket_sweep_main} helps interpret PaMM as a bounded-capacity memory rather than as a free-form feature branch. Increasing the number of pair/triplet buckets from 2k to the 8k regime improves both validation and test metrics, while the move from 8k to 16k yields only marginal additional energy benefit and no consistent force improvement. This diminishing-return pattern is what one would expect if collisions matter at very small memory sizes but the useful recurring motifs are already covered once the table reaches a moderate capacity. The sweep therefore supports the claim that PaMM's added state is doing organized work, not merely absorbing more parameters without structure.

\paragraph{Scope.}
The current evidence supports a narrow story: PaMM is a useful local-structure inductive bias in UMA-S on OMAT, especially at the matched 10k budget, and its benefit is tied to structured pair/triplet organization. The appendix contains the broader but less acceptance-critical evidence package: mechanism-oriented plots, compute and reproducibility notes, a mixed Alexandria-PBE comparability check, a tracked medium-host placeholder, a planned MPtrj placeholder, and a harder leave-one-family-out pilot. We include those boundary checks to mark the present empirical limits of the method rather than to broaden the central claim beyond the matched OMAT comparison above.

\section{Conclusion}
\label{sec:conclusion}

PaMM augments the UMA eSCN-MD edge encoder with explicit pair and triplet motif memory while leaving the downstream equivariant backbone unchanged. In the studied UMA-S + OMAT setting, this explicit memory improves the plain baseline at a matched 10k budget and remains competitive at 20k, with gate-only and affine-equipped variants favoring different metrics. The main-text analyses further show that the effect is visible throughout the relevant training horizon, remains directionally stable under reseeding, weakens under aligned structure-destroying controls, and exhibits diminishing returns as the memory capacity grows. Taken together, these results support a focused claim: explicit motif memory is a useful local-structure inductive bias for this host/domain/budget regime.

\paragraph{Limitations.}
The evidence is strongest in the UMA-S + OMAT setting with matched 10k/20k budgets, and the implemented recipe couples structured memory with the reduced lookup/gate learning-rate group used in code. Appendix boundary checks are mixed: the sampled Alexandria-PBE comparison is close but slightly unfavorable to PaMM, the leave-one-family-out \texttt{rattled} pilot is only mildly positive, and the tracked medium-host and planned MPtrj comparisons were not yet finalized at paper freeze. We therefore do not claim broad cross-dataset transfer or host-scaling gains. We also do not claim strong chemistry-level interpretability beyond a more inspectable local-structure interface. These questions, together with higher-order motif memory and fuller resource benchmarking, remain open.

\clearpage
\bibliography{references}
\bibliographystyle{plainnat}

\clearpage
\appendix
\section{Appendix}

\subsection{Reproducibility, Compute Resources, Limitations, and Broader Impact}
\label{sec:appendix_repro}

\paragraph{Reproducibility details.}
All reported main-text experiments use the UMA-S-1p1 eSCN-MD host, the same local OMAT split of \texttt{dataset=uma}, the same training entrypoint, and matched optimization settings. Unless otherwise noted, PaMM uses separate pair and triplet tables, two hash functions, discretized distance and angle bins, and a reduced learning-rate group for the lookup and gate submodules. The main-text tables report locked 10k and 20k checkpoints exported from the same training workflow; the control rows use the same host and checkpoint protocol with only the stated mechanism change.

\paragraph{Compute resources.}
Paper-critical training runs used the local \texttt{h100\_local} CUDA profile with two ranks on one node, 24 CPU cores per task, no dataloader workers, bf16 training, and a max-atom cap of 3000 per rank. Representative launch logs for the baseline and PaMM runs show steady-state training throughput on the order of $1.4\text{k}$--$1.9\text{k}$ atoms/s per rank during ordinary train intervals. Checkpoint-time ETA logs around the long baseline and PaMM runs place the remaining 10k-step half in the 7.7--7.9 hour range, so the locked 10k and 20k checkpoints correspond roughly to 7--8 hour and 15--16 hour jobs under this profile. The full project consumed more compute than the paper-critical runs because earlier tuning waves, negative controls, and discarded variants are not counted in the main tables.

\paragraph{Memory footprint and deployment.}
PaMM's additional state is concentrated in the hash lookup tables. At the paper-default setting from the released config family ($d=256$, $H=2$, $M_{\mathrm{pair}}=M_{\mathrm{tri}}=8192$), the pair and triplet tables contain $8{,}388{,}608$ scalars in total, corresponding to about 16 MB in bf16 or 32 MB in fp32. Because this state is accessed through sparse integer-indexed lookup rather than through dense equivariant activations, it is amenable to host-memory or managed-memory placement at deployment, which can make its marginal VRAM cost close to zero relative to the GPU-resident backbone. The present training implementation keeps the tables in the standard model module for simplicity, so the paper does not claim a separate memory-offload benchmark.

\paragraph{Scope and limitations.}
The paper's strongest evidence is host- and regime-specific: UMA-S on OMAT under matched 10k/20k budgets. The implementation also evaluates the complete PaMM recipe available in code, so structured memory is not cleanly separated from the reduced lookup/gate learning-rate group. In addition, PaMM uses discretized geometry and fixed-size hash tables, so information loss and collisions are reduced rather than eliminated, and the present work stops at pair/triplet structure instead of higher-order motif memory.

\paragraph{Broader impact.}
PaMM is a foundational atomistic-modeling method aimed at more data-efficient and inspectable prediction for molecules and materials. A plausible positive impact is faster scientific screening in catalysis and materials discovery. Potential negative impact is limited but not zero: better atomistic surrogate models can also accelerate screening of harmful or environmentally problematic compounds. We therefore view PaMM as a research contribution that should be released with the same caution and benchmark-context framing already used for existing atomistic-model assets.

\subsection{Detailed Fusion Equations and Implementation Notes}

\paragraph{Edge-wise gated fusion.}
Given the augmented edge feature $\edgefeat_{ij}$, the gate-only variant computes
\begin{align}
q_{ij} &= W_q \edgefeat_{ij}, \\
k_{ij} &= W_k \edgefeat_{ij}, \\
s_{ij} &= \langle q_{ij}, k_{ij} \rangle, \\
\hat{s}_{ij} &= \mathrm{clip}(\alpha s_{ij}, -c, c), \\
a_{ij} &= \mathrm{SiLU}(\hat{s}_{ij}), \\
g_{ij} &= \mathrm{clip}(1 + \lambda a_{ij}, g_{\min}, g_{\max}), \\
\tilde{\edgefeat}_{ij} &= \edgefeat_{ij} \odot g_{ij}.
\end{align}
This is an edge-local scalar gate rather than a multi-edge attention mechanism.

\paragraph{Affine-equipped fusion.}
The affine-equipped variant uses the same memory features but also injects them inside each message-passing layer. Let
\begin{equation}
m_{ij} = [\pairmem_{ij}; \tripmem_{ij}].
\end{equation}
For each layer $\ell$, a small MLP produces
\begin{align}
c^{(\ell)}_{ij} &= \mathrm{MLP}_{\ell}(m_{ij}), \\
\Delta g^{(\ell)}_{ij} &= W^{(\ell)}_{g} c^{(\ell)}_{ij}, \\
\Delta b^{(\ell)}_{ij} &= W^{(\ell)}_{b} c^{(\ell)}_{ij}.
\end{align}
The implemented affine coefficients are
\begin{align}
g^{(\ell)}_{ij} &= \alpha_{\mathrm{aff}} \tanh(\Delta g^{(\ell)}_{ij}), \\
b^{(\ell)}_{ij} &= \beta_{\mathrm{aff}} \Delta b^{(\ell)}_{ij},
\end{align}
and the layer-specific edge state is updated as
\begin{equation}
e^{(\ell)}_{ij} \leftarrow e^{(\ell)}_{ij} \odot (1 + g^{(\ell)}_{ij}) + b^{(\ell)}_{ij}.
\end{equation}
The affine output heads are zero-initialized so that the branch starts as a small correction path.

\paragraph{Variant-to-flag mapping.}
The implementation exposes the method family through a small set of toggles:
\begin{itemize}
\item \textbf{Baseline}: disable PaMM lookup entirely.
\item \textbf{Structured PaMM}: enable pair/triplet memory and scalar gating.
\item \textbf{Pair-only / triplet-only}: disable one motif source while keeping the other.
\item \textbf{No-gate}: keep structured pair/triplet lookup but bypass scalar edge gating.
\item \textbf{Random-bucket}: keep lookup-style capacity while replacing structured keys with unstructured bucket assignment.
\item \textbf{MLP control}: replace structured lookup features by a parameter-matched generic MLP branch.
\item \textbf{Affine modulation}: keep the structured gated path and add per-layer channel-wise affine modulation driven by the same memory features.
\item \textbf{Reduced lookup/gate LR}: place the lookup and gate submodules into a dedicated optimizer group with a smaller learning-rate scale than the host backbone.
\end{itemize}

\subsection{Companion-Wave Robustness Results}

We also completed a companion wave designed to stress the paper's central comparison without changing its scope. Its role is narrower than the main-text tables: it asks whether the locked 10k-step comparison remains directionally stable across repeated seeds, and how the same variants compare under one moderately longer 20k-step budget.

\begin{table*}[t]
\centering
\caption{\textbf{Companion-wave summary for repeated 10k runs and matched 20k follow-up.} For the 10k protocol, we report mean $\pm$ std over three seeds. The 20k follow-up was run once per variant under the same host and training recipe. Lower is better for all MAEs.}
\label{tab:pending_companion_wave}
\scriptsize
\setlength{\tabcolsep}{3.5pt}
\begin{tabular}{lcccccccc}
\toprule
& \multicolumn{4}{c}{10k protocol} & \multicolumn{4}{c}{20k follow-up} \\
\cmidrule(lr){2-5} \cmidrule(lr){6-9}
Variant & Val E & Val F & Test E & Test F & Val E & Val F & Test E & Test F \\
\midrule
Plain baseline
& 0.3229$\pm$0.0012
& 0.0951$\pm$0.0004
& 0.4467$\pm$0.0094
& 0.0984$\pm$0.0005
& 0.2667
& 0.0830
& 0.3866
& 0.0869 \\
Gate-only PaMM
& \textbf{0.3193}$\pm$0.0026
& \textbf{0.0951}$\pm$0.0001
& \textbf{0.4440}$\pm$0.0037
& 0.0984$\pm$0.0001
& \textbf{0.2638}
& 0.0829
& 0.3808
& 0.0867 \\
Affine-equipped PaMM
& 0.3243$\pm$0.0018
& 0.0953$\pm$0.0004
& 0.4474$\pm$0.0069
& 0.0984$\pm$0.0004
& 0.2696
& \textbf{0.0828}
& \textbf{0.3757}
& \textbf{0.0865} \\
\bottomrule
\end{tabular}
\end{table*}

The repeated 10k comparison preserves the energy-oriented gate-only advantage on average, but the force differences remain extremely small. The 20k follow-up keeps the same main interpretation as the paper: the memory mechanism remains competitive, while the fusion choice mainly shifts the operating point rather than revealing a uniformly dominant late-budget winner.

\subsection{Anytime and Source-Family Visualizations}

\begin{figure*}[t]
\centering
\includegraphics[width=0.88\linewidth]{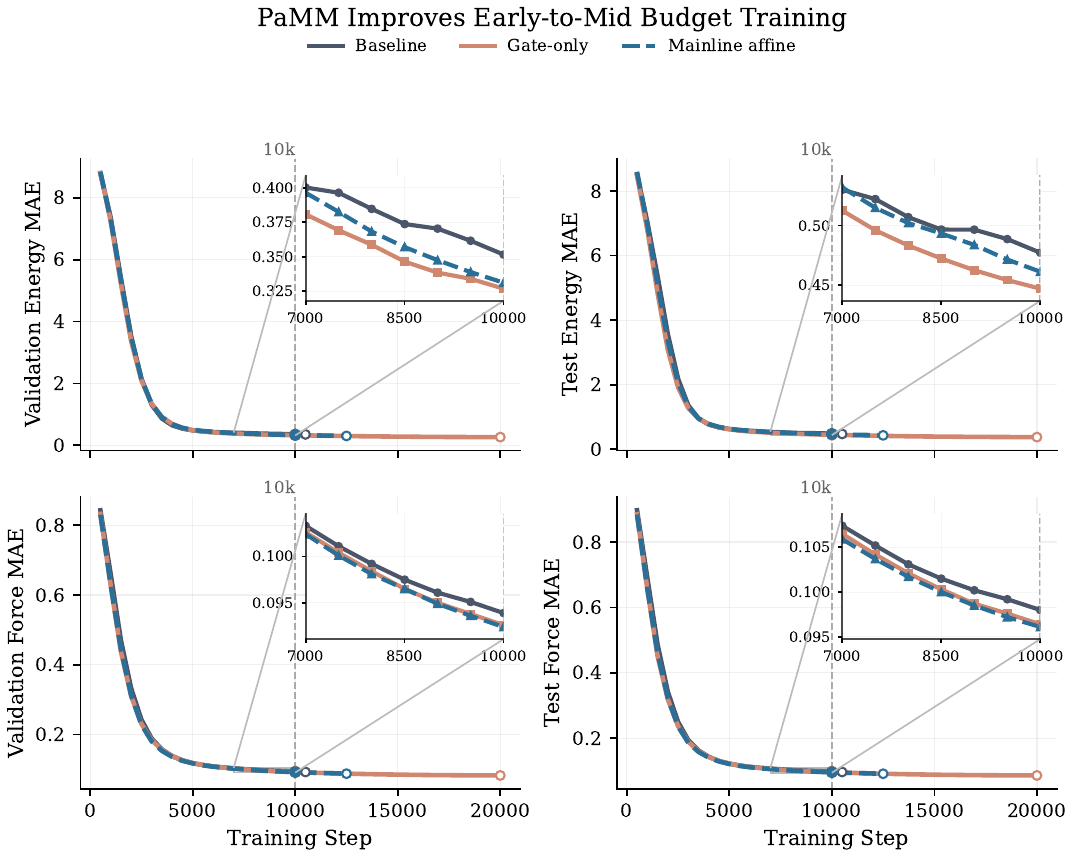}
\caption{\textbf{Anytime validation and test curves under the matched protocol.} Both PaMM variants improve on the plain UMA-S baseline before full convergence. Gate-only is strongest on energy near the locked budget, while the affine-equipped variant is strongest on force.}
\label{fig:main_anytime}
\end{figure*}

\begin{figure*}[t]
\centering
\includegraphics[width=0.88\linewidth]{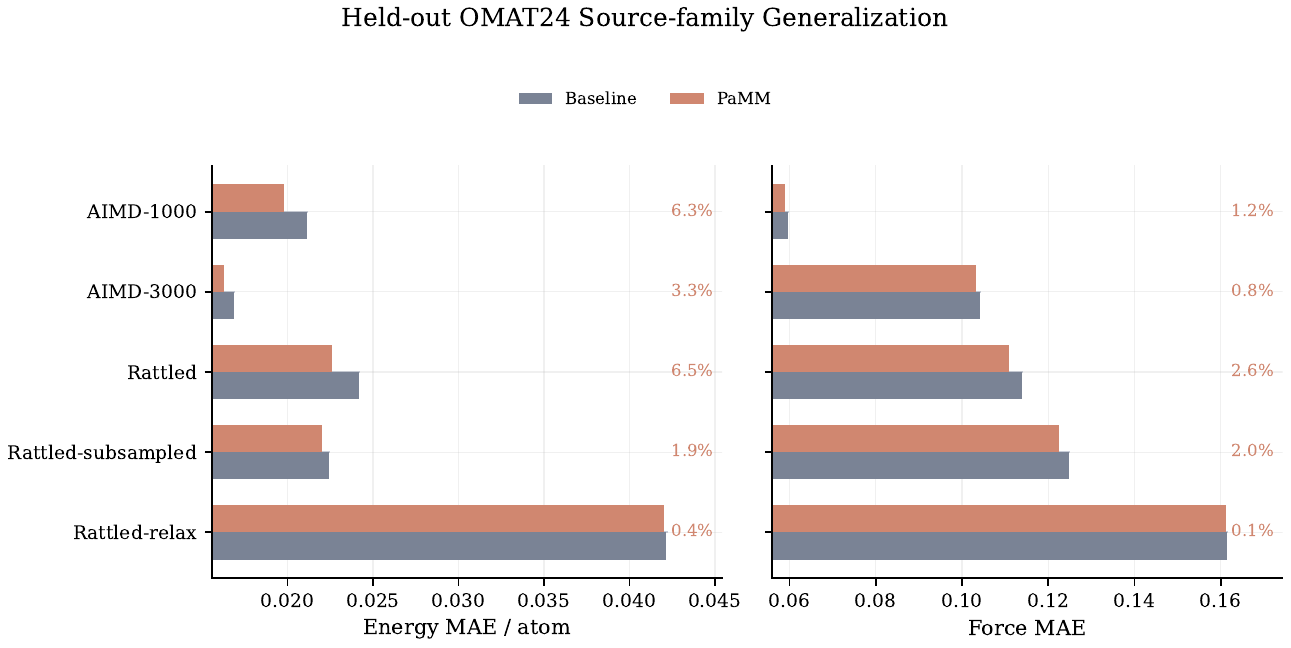}
\caption{\textbf{Held-out OMAT24 source-family comparison at the matched checkpoint.} Bars compare the plain baseline and the evaluated PaMM checkpoint on each source family. The PaMM checkpoint is consistently better across all five source families on both energy and force MAE.}
\label{fig:source_family}
\end{figure*}

\subsection{Visual Summary of the Controls}

\begin{figure}[t]
\centering
\includegraphics[width=0.95\linewidth]{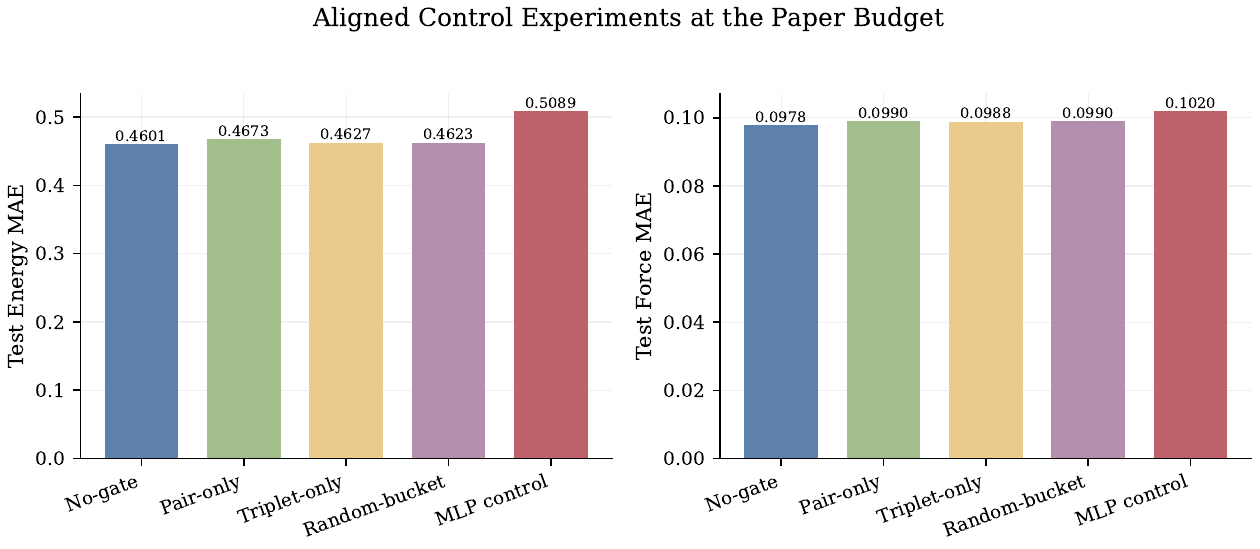}
\caption{\textbf{Visual summary of the aligned 10k-step controls.} Pair-only and triplet-only remain weaker than the full pair+triplet memory path, while the parameter-matched MLP control is the clearest negative reference.}
\label{fig:controls_appendix}
\end{figure}

\subsection{External and OOD Boundary Checks}

The appendix is the right place for external evidence that is informative but less tightly protocol-matched than the main OMAT comparison. The role of these experiments is not to upgrade the paper into a broad transfer claim, but to check how the proposed memory path behaves once evaluation is pushed outside the main in-domain setting. The picture is mixed rather than decisive: the sampled Alexandria-PBE comparison is slightly unfavorable to PaMM, the harder family-holdout pilot remains only mildly positive, and the planned MPtrj benchmark was not yet run at paper freeze. We therefore treat all three as boundary evidence rather than as core support for the paper's main claim.

\begin{table}[t]
\centering
\small
\begin{tabular}{lcccc}
\toprule
\textbf{Variant} & \textbf{Shards} & \textbf{Samples} & \textbf{Energy / atom} $\downarrow$ & \textbf{Force} $\downarrow$ \\
\midrule
Baseline & 10 & 5000 & \textbf{0.05799} & \textbf{0.02561} \\
PaMM & 10 & 5000 & 0.05853 & 0.02588 \\
\midrule
Baseline & 20 & 10000 & \textbf{0.06076} & \textbf{0.02535} \\
PaMM & 20 & 10000 & 0.06122 & 0.02552 \\
\bottomrule
\end{tabular}
\caption{\textbf{Alexandria-PBE external benchmark at the 10k training checkpoint.} We report two evaluation sample counts from stored benchmark logs. PaMM remains close to the plain baseline, but in this sampled external comparison it is slightly worse on both energy and force MAE at both sizes. We therefore treat Alexandria as a comparability check rather than as positive transfer evidence for the main paper claim.}
\label{tab:alexandria_external}
\end{table}

\subsection{Tracked Medium-Host Boundary Check}

The experiment record also includes a planned host-scaling check beyond UMA-S. In the tracked wave, \texttt{R024} and \texttt{R025} correspond to a medium-host OMAT comparison with a \texttt{K10L4} setting and \texttt{max\_atoms=700}, intended as appendix-only boundary evidence rather than as a main-paper rescue path. At the paper-freeze snapshot used for this submission, both tracker entries were still marked as running and no stable checkpoint-level evaluation summary had been archived. We therefore record the intended setup below without using it as numerical evidence.

\begin{table}[t]
\centering
\small
\begin{tabular}{llll}
\toprule
\textbf{Tracker ID} & \textbf{Host / setting} & \textbf{Variant} & \textbf{Status} \\
\midrule
\texttt{R024} & Medium host (\texttt{K10L4}, \texttt{max\_atoms=700}) & Baseline & \texttt{RUNNING} \\
\texttt{R025} & Medium host (\texttt{K10L4}, \texttt{max\_atoms=700}) & PaMM & \texttt{RUNNING} \\
\bottomrule
\end{tabular}
\caption{\textbf{Experiment-record placeholder for the planned medium-host OMAT boundary check.} The tracker contains the intended baseline and PaMM runs beyond UMA-S, but no stable evaluation summary was archived at paper freeze. We therefore report only the tracked setup and status here, not numerical results.}
\label{tab:medium_host_tracker}
\end{table}

\subsection{Planned MPtrj External Placeholder}

We also reserve MPtrj as a planned external benchmark beyond the current OMAT-centered study. This comparison was not yet run at paper freeze, so we keep it only as a placeholder to make the intended evidence package explicit rather than to imply that an unreported result was favorable.

\begin{table}[t]
\centering
\small
\begin{tabular}{llll}
\toprule
\textbf{Dataset} & \textbf{Compared systems} & \textbf{Intended budget} & \textbf{Status} \\
\midrule
\texttt{MPtrj} & Baseline vs.\ PaMM & Locked 10k checkpoint export & \texttt{NOT\_RUN} \\
\bottomrule
\end{tabular}
\caption{\textbf{Placeholder for planned MPtrj external evaluation.} No stable MPtrj run or archived metric summary was available at paper freeze, so the paper does not use this benchmark as evidence. We include the placeholder only to document the intended next external check.}
\label{tab:mptrj_placeholder}
\end{table}

\subsection{Harder OOD Family-Holdout Pilot}

We also ran a first harder OOD pilot under a leave-one-source-family-out protocol. The completed pilot was a \texttt{rattled} holdout: all \texttt{rattled-300/500/1000} training sources were removed from the training mix, and evaluation was performed only on held-out \texttt{rattled} validation and test splits. We compare only the plain baseline and the gate-only PaMM variant, keeping the same UMA-S host and the same 10k-step budget as in the main paper.

\begin{table}[t]
\centering
\small
\begin{tabular}{lcccc}
\toprule
\textbf{Variant} & \textbf{Val E} $\downarrow$ & \textbf{Val F} $\downarrow$ & \textbf{Test E} $\downarrow$ & \textbf{Test F} $\downarrow$ \\
\midrule
Gate-only PaMM & \textbf{0.3611} & \textbf{0.1498} & \textbf{0.4719} & 0.1485 \\
Baseline & 0.3627 & 0.1499 & 0.4803 & 0.1485 \\
\bottomrule
\end{tabular}
\caption{\textbf{Leave-one-\texttt{rattled}-family-out OOD pilot at the 10k-step budget.} Gate-only PaMM is slightly better on validation energy/force and test energy, while test force is tied. We therefore treat this as a boundary check rather than as core supporting evidence.}
\label{tab:rattled_holdout_pilot}
\end{table}

This pilot is informative because it is only mildly positive rather than decisive. Under the harder family-holdout shift, PaMM remains broadly comparable to the baseline and does not yet justify a stronger OOD headline. Its value is to mark the current boundary of the paper's strongest claim, especially when read together with the mixed Alexandria comparison and the currently unfilled MPtrj placeholder above.

\subsection{Mechanism Analyses}

The following plots are intended as mechanism-oriented consistency checks rather than as causal proof. They ask whether the observed gains align with the proposed story that periodic materials repeatedly instantiate motifs that can be reused through an explicit memory path.

\begin{figure}[t]
\centering
\includegraphics[width=0.82\linewidth]{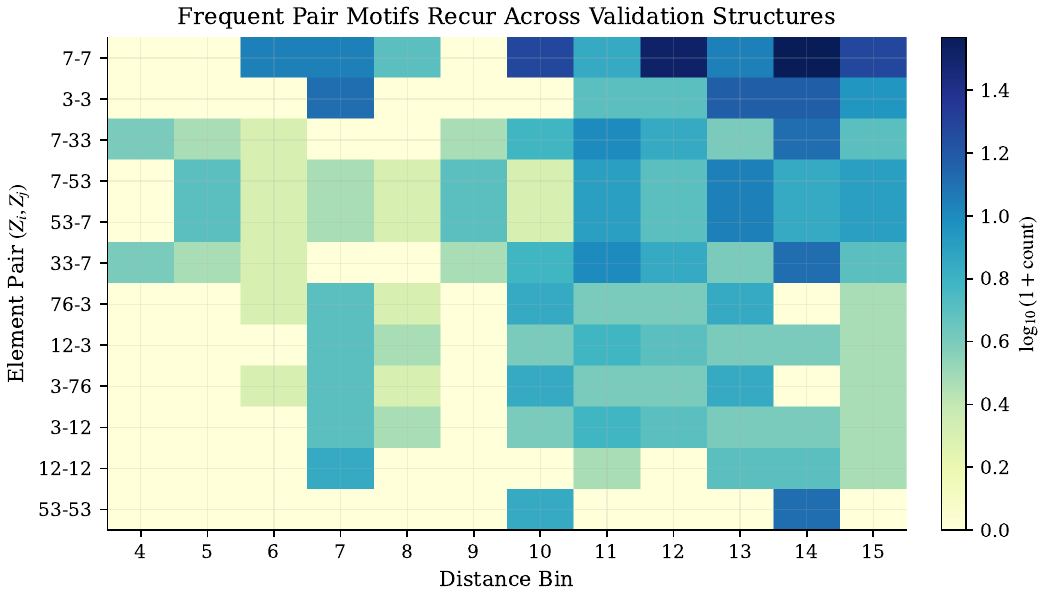}
\caption{\textbf{Frequent pair motifs recur across the analyzed OMAT24 validation slice.} A relatively small subset of pair types occupies a large fraction of the motif mass.}
\label{fig:pair_motif_heatmap}
\end{figure}

\begin{figure}[t]
\centering
\includegraphics[width=0.78\linewidth]{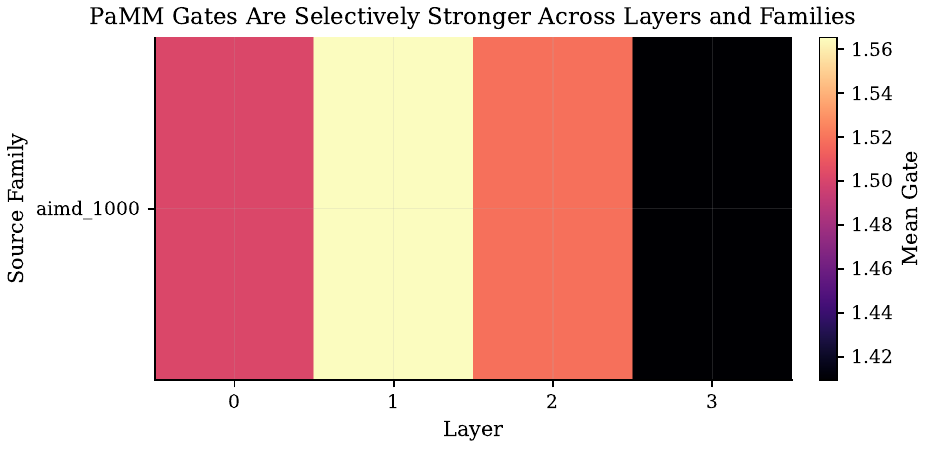}
\caption{\textbf{Gate usage is structured across layers and source families.} The mean scalar gate magnitude is not uniform, suggesting that the memory path is used selectively rather than as a static extra feature block.}
\label{fig:gate_heatmap}
\end{figure}

\begin{figure}[t]
\centering
\includegraphics[width=0.88\linewidth]{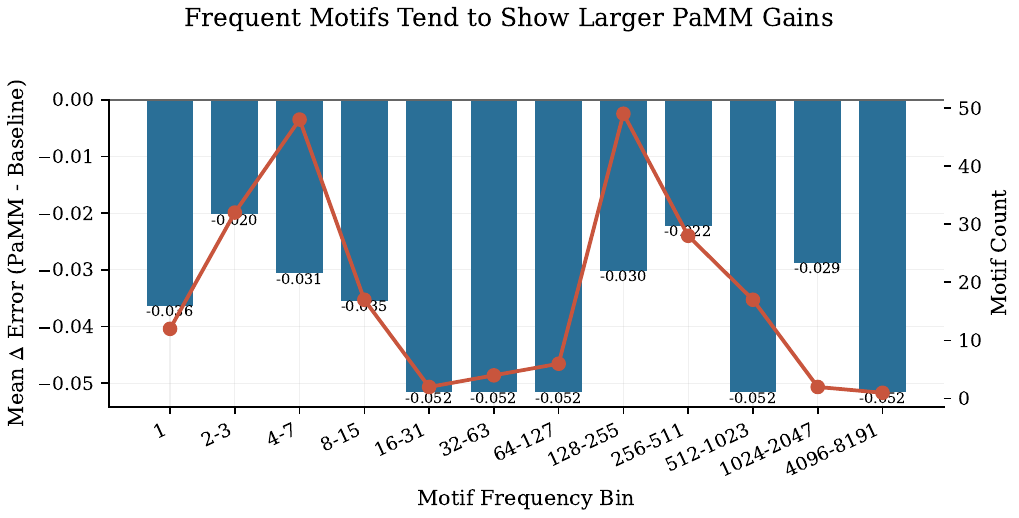}
\caption{\textbf{More frequent motifs tend to align with larger PaMM gains in the analysis slice.} The plotted quantity is the mean error change $(\mathrm{PaMM} - \mathrm{baseline})$, so lower is better.}
\label{fig:motif_effect_by_frequency}
\end{figure}

\Cref{fig:pair_motif_heatmap} shows that the empirical motif distribution is concentrated rather than flat, which supports the premise that periodic local geometry presents repeated objects worth storing. \Cref{fig:gate_heatmap} shows that the memory path is used selectively across layers and source families rather than behaving like a uniform feature concatenation. \Cref{fig:motif_effect_by_frequency} is correlational, but it is directionally consistent with the intended reuse story: more frequent motifs tend to align with more favorable PaMM--baseline deltas.

\subsection{Theory-Only Overhead Analysis}

\begin{table}[t]
\centering
\caption{\textbf{Theoretical overhead decomposition of PaMM.} Here $E$ is the number of directed edges, $T$ is the number of enumerated triplets, $d$ is the memory width, $H$ is the number of hash functions, and $L$ is the number of message-passing layers.}
\label{tab:theory}
\small
\begin{tabular}{p{0.31\linewidth}p{0.60\linewidth}}
\toprule
Component & Added cost \\
\midrule
Pair hash memory & parameters $\mathcal{O}(H M_{\mathrm{pair}} d)$; lookup work linear in $E$ \\
Triplet hash memory & parameters $\mathcal{O}(H M_{\mathrm{tri}} d)$; enumeration and aggregation linear in $T$ \\
Edge-wise gate & edge-local projections and scalar modulation, linear in $L$ and active edges \\
Affine modulation & per-layer memory MLP plus channel-wise affine update, linear in $L$, $E$, and edge width \\
\bottomrule
\end{tabular}
\end{table}

PaMM does not introduce a second message-passing stack or a new all-to-all interaction pattern. Its added work is local lookup plus local modulation, which is why we view it as a controlled additive memory path rather than a replacement backbone.

\subsection{Bucket-Count Sweep}

We also swept the PaMM bucket count on UMA-S under the same 10k-step protocol, using matched pair and triplet table sizes of 2048, 4096, 8192, and 16384. The trend is stable: increasing the bucket count from 2k to 8k improves both validation and test metrics, while moving from 8k to 16k gives only marginal additional energy benefit and no consistent force improvement.

\begin{table}[t]
\centering
\small
\begin{tabular}{lcccc}
\toprule
\textbf{Buckets} & \textbf{Val E} $\downarrow$ & \textbf{Val F} $\downarrow$ & \textbf{Test E} $\downarrow$ & \textbf{Test F} $\downarrow$ \\
\midrule
2048 & 0.04435 & 0.10507 & 0.06121 & 0.10851 \\
4096 & 0.04373 & 0.10515 & 0.06009 & 0.10844 \\
8192 & \textbf{0.04301} & \textbf{0.10464} & 0.05905 & \textbf{0.10833} \\
16384 & 0.04315 & 0.10487 & \textbf{0.05818} & 0.10849 \\
\bottomrule
\end{tabular}
\caption{\textbf{Bucket-count sweep for PaMM on UMA-S at the 10k checkpoint.} Performance improves substantially from 2k to the 8k--16k regime, with diminishing returns beyond 8k.}
\label{tab:bucket_sweep}
\end{table}

\begin{figure}[t]
\centering
\includegraphics[width=0.85\linewidth]{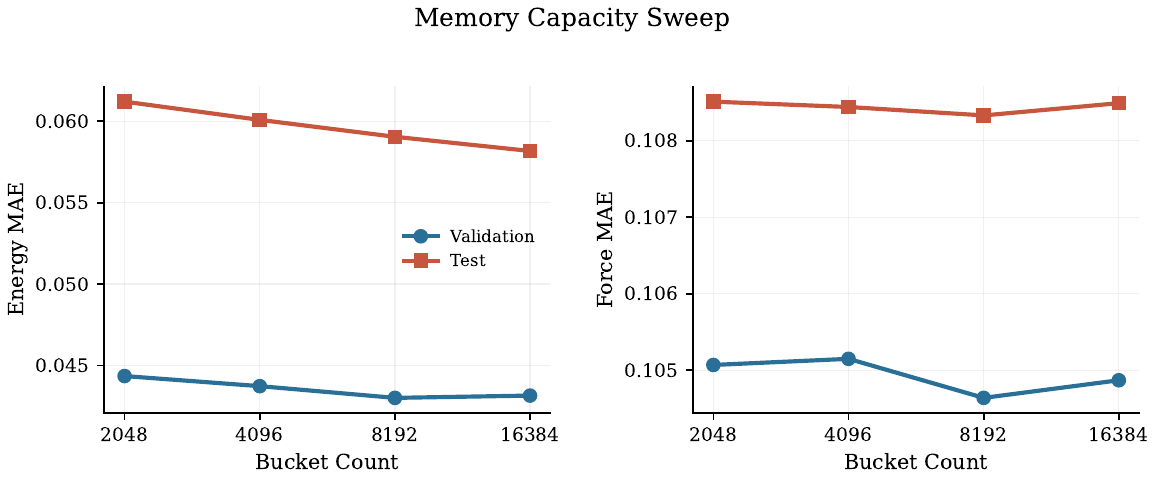}
\caption{\textbf{Bucket-count sweep at the matched 10k-step budget.} Increasing the number of pair/triplet memory buckets from 2k to 8k improves both validation and test metrics, while the move from 8k to 16k gives only marginal additional benefit.}
\label{fig:bucket_sweep}
\end{figure}

\end{document}